\newcommand\Mark[1]{\textsuperscript#1}

\documentclass{sig-alternate}

\usepackage[
  pass,
]{geometry}

\newfont{\mycrnotice}{ptmr8t at 7pt}
\newfont{\myconfname}{ptmri8t at 7pt}
\let\crnotice\mycrnotice%
\let\confname\myconfname%

\permission{This is a preprint version prior to submission for peer-review.}
\conferenceinfo{ACM MultiMedia'14,}{November 3--7, 2014, Orlando, Florida, USA.}
\copyrightetc{Creative Commons Attribution-NonCommerical 4.0 International \the\acmcopyr}
\crdata{DOI for the accepted paper: 
http://dx.doi.org/10.1145/2647868.2654896 }

\usepackage{graphicx}
\usepackage{microtype}
\usepackage{booktabs}
\usepackage{amsmath}
\usepackage{amssymb}
\usepackage{listings}
\usepackage{courier}
\usepackage{url}
\usepackage{color}
\usepackage{paralist}

\def\sharedaffiliation{%
\end{tabular}
\begin{tabular}{c c}}
\begin{document}
%

\title{Object Segmentation in Images using EEG Signals}

%
%
%
%
%

\numberofauthors{6} 
%
\author{
%
%
\alignauthor
Eva Mohedano\Mark{1}\\
       \email{eva.mohedano@insight-centre.org}
\alignauthor
Graham Healy\Mark{1}\\
       \email{ghealy@computing.dcu.ie}
\alignauthor Kevin McGuinness\Mark{1}\\
       \email{kevin.mcguinness@dcu.ie}
\and  
\alignauthor Xavier Gir{\'o}-i-Nieto\Mark{2}\\
       \email{xavier.giro@upc.edu}
\alignauthor Noel E. O'Connor\Mark{1}\\
       \email{noel.oconnor@dcu.ie}
\alignauthor Alan F. Smeaton\Mark{1}\\
       \email{alan.smeaton@dcu.ie}
\sharedaffiliation
	     \affaddr{\Mark{1}Insight Centre for Data Analytics} & \affaddr{\Mark{2}Image Processing Group}\\
        \affaddr{Dublin City University} & \affaddr{Universitat Polit{\`e}cnica de Catalunya}\\
         \affaddr{Glasnevin, Dublin 9, Ireland}& \affaddr{C. Jordi Girona, 1-3. 08034 Barcelona}
}

\additionalauthors{Additional authors: John Smith (The Th{\o}rv{\"a}ld Group,
email: {\texttt{jsmith@affiliation.org}}) and Julius P.~Kumquat
(The Kumquat Consortium, email: {\texttt{jpkumquat@consortium.net}}).}
\date{19 Mar 2014}

\maketitle
\begin{abstract}
This paper explores the potential of brain-computer interfaces in segmenting objects from images. Our approach is centered around designing an effective method for displaying the image parts to the users such that they generate measurable brain reactions. When an image region, specifically a block of pixels, is displayed we estimate the probability of the block containing the object of interest using a score based on EEG activity. After several such blocks are displayed, the resulting probability map is binarized and combined with the GrabCut algorithm to segment the image into object and background regions. This study shows that BCI and simple EEG analysis are useful in locating object boundaries in images.
\end{abstract}

\category {I.4.6} Segmentation



\section{Introduction}
\label{sec:intro}

The human brain is capable of processing audiovisual information in a fashion that, nowadays, clearly outperforms machines in several applications. 
The multimedia research community is constantly trying to simulate the brain's behaviour to later leverage its innate computational possibilities through machinery.
However, a deep understanding of the human brain remains one of the greatest scientific challenges.
Recent initiatives, such us the Human Brain Project in Europe or the the BRAIN Initiative in the United States, have identified its exploration as one of the Grand Challenges of our time. 

Although humans consistently outperform computers in the semantic interpretation of multimedia signals \cite{hu2012bridging}, the computational and storage power of machines can be scaled and networked dramatically beyond individual human capacities.
These two observations are the foundation of the human computational technologies, which exploit the best of both by defining collaborative strategies.
The steady decrease in the cost of EEG (Electroencephalography) systems in recent years has made these non-invasive Brain-Computer Interfaces (BCIs) accessible beyond the traditional disciplines that typically availed of this technology~\cite{Zhong-HumanComputation-2009,sajda2010blink}.
Visual analysis is one such field, with recent publications exploring the potential of EEG signals for image retrieval~\cite{Healy:2011,conf:icip:YazdaniVIAE10,Wang:2009} and object detection~\cite{Bigdely-Shamlo:2008,kapoor2008combining}.

The use of brain-computer interfaces is, however, still limited, primarily because the motor (or speech) capabilities of most humans provide richer interaction methods than BCIs.
For this reason, many current applications use BCIs as a secondary interaction source to complement another primary one, or as a tool for scientists to study human behaviour~\cite{Hebbalaguppe2013}. Brain-computer interfaces, however, have the potential to be enormously beneficial for seriously impaired people, such as those affected by \emph{Locked In Syndrome} (LIS). These individuals are paralysed of nearly all voluntary muscles, so are disabled from motion and speech. Vision is always intact, although in extreme cases even eye movement is restricted~\cite{Bauer:LIS-NoEye:1979}, in which cases BCIs represent the only opportunity to interact with the world.

Although a controversial discussion topic between neuroscientists, some authors claim to have observed consciousness with EEG devices on patients with persistent vegetative state~\cite{cruse2012bedside}, which may open a door to a certain interaction with them.
For these reasons, and as explained in~\cite{Fernandez-MSc-2013}, \emph{BCI systems hold great promise for effective basic communication capabilities through machines, e.g. by controlling a spelling program or operating a neuroprosthesis}.
The use of EEGs for these type of assistive technologies has been previously explored in applications like letter-by-letter spelling~\cite{Roark:RSVPkeyboard:2013} or the control of robots~\cite{Bell:HumanoidRobot:2008,Pathirage:wheelchair:2013}.


The objective of this work is to demonstrate that BCI interfaces are useful in tasks beyond spelling out words. We focus here on interaction with multimedia: specifically, object selection and segmentation in images. The capacity to perform such segmentation using a BCI interface potentially has both practical and creative applications, such as selection of specific objects for similarity search, and mixing objects from different sources to create a new composition. We propose a system capable of accurately selecting an object in an image in a manner that is completely hands-free, using only measured signals from an EEG interface. In this way, previous work exploring image retrieval (global image scale)~\cite{Healy:2011,conf:icip:YazdaniVIAE10,Wang:2009} and object detection (coarse local scale)~\cite{Bigdely-Shamlo:2008,kapoor2008combining} are extended to a pixel-level object segmentation. This task is addressed by applying the human computation paradigm, using noisy EEG signals to seed the well-known GrabCut~\cite{Rother:2004:GIF:1015706.1015720} segmentation algorithm.

This remainder of the paper is structured as follows. Section~\ref{sec:relatedWork} reviews previous work exploring the use of EEG signals for multimedia analysis.
Section~\ref{sec:systemArchitecture} provides an overview of the entire system architecture, which is described in detail in Sections~\ref{sec:dataacquisition},~\ref{sec:eegprocessing}, and~\ref{sec:segmentation}. 
Section~\ref{sec:results} presents the results from out experiments.
Section~\ref{sec:conclusionsandfuturework} gives conclusions and outlines future research directions.

\section{Related Work}
\label{sec:relatedWork}
Previous work combining BCI and computer vision~\cite{Healy:2011,Wang:2009,Huang:2011} have been focused primarily on image retrieval and object detection. In such work images are presented to participants according to the \textit{oddball paradigm}. This approach consists of presenting a ``target'' image among many ``distractor'' images via Rapid Serial Visual Presentation (RSVP)~\cite{journals/ivs/Spence02}. The presentation rate of the images is high, around 10Hz, so that a specific signature in the corresponding EEG signals is produced when the user observes the target images (or rare stimulus). This signature is known as a P300 wave and it is a kind of Event-Related Potential (ERP) associated to the process of recognising a relevant visual stimulus~\cite{luck2005introduction}. The wave's primary characteristic is a positive peak in the EEG signal 300ms after the visual stimulus was observed.

Two previous works describing a BCI system applied to image retrieval and detection were presented by Wang~\cite{Wang:2009} and Healy~\cite{Healy:2011}. In both cases the authors perform RSVP of images from known datasets 
at 10Hz to detect those images in which a specific object appears. The main difference between them is that in Wang's paper the user is not asked to press any additional button when a target image is seen. Our work differs from these because it focuses on target windows (or regions) instead of target image detection. The most similar work to ours is Bigdely-Shamlo's paper~\cite{Bigdely-Shamlo:2008}, in which satellite images are explored using local windows to detect those containing airplanes. Bigdely-Shamlo's work, however, assumes that the object fits in a single window, while in our contribution objects are partially represented in an unknown number of windows.

\section{System Architecture}
\label{sec:systemArchitecture}

We propose a system that aims to both detect and segment an object from an image using the measured brain signals of the user at the moment of observing a specific region. 
The idea is to transform the measured EEG responses into a map that gives an estimate of how probable it is that a particular region seen by the user contains the target object, and then use this map to seed a segmentation algorithm.
The construction of this map is based on EEG signal classification, as the electrical responses of the brain are known to differ when the user detects a target or rare stimulus in a RSVP scenario.

\begin{figure}[t]
\centering
\includegraphics[width=1.0\columnwidth]{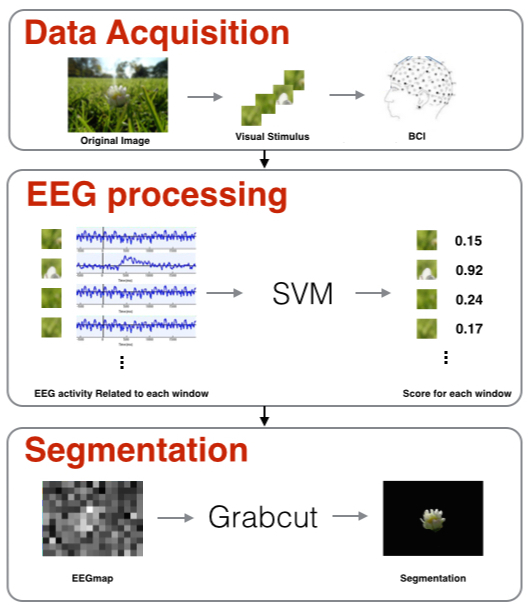}
\caption{Block diagram of the system.
}
\label{fig:systemArchitecture}
\end{figure}

Figure~\ref{fig:systemArchitecture} illustrates the three primary stages of the proposed system:
\begin{enumerate}
\item \textbf{Data acquisition} (Section~\ref{sec:dataacquisition}): in this stage we capture the brain signals related to the visual stimulus.
\item \textbf{EEG processing} (Section~\ref{sec:eegprocessing}): pre-processing and classification are used to generate the probability maps for the object location. As these maps are built by using EEG analysis, they will be referred to as~\textit{EEG maps}.
\item \textbf{Segmentation} (Section~\ref{sec:segmentation}): EEG maps are used to seed the GrabCut object segmentation algorithm~\cite{Rother:2004:GIF:1015706.1015720}.
\end{enumerate}
The following sections of the paper describe each stage in more detail.

\section{Data acquisition}
\label{sec:dataacquisition}

This section describes the experimental set-up used to capture the data.
First, a new image dataset was created and each image partitioned in blocks of equal size.
Each of these blocks are presented at a high rate, in order to generate a measurable response on EEG signals.
This stage was validated with a preliminary test with a single user, an important step before starting a larger campaign of data acquisition.
After the positive output from the preliminary test, the final experiments reported in the remainder of the paper were based on a population of five people between 21 and 32 years old.

\subsection{Image dataset}
\label{ImageDataset}

A novel dataset of 22 images was created to run the experimentation described in this paper.
Given the exploratory nature of this work, the images were chosen to include a single object in a background of limited complexity.
The dataset includes different configurations regarding the color, shape, and texture of the objects, as well as their relative similarity with the foreground.

The collection consists of 20 new images captured for the purpose of this work and images 38082 and 123074 from the \emph{Berkeley Segmentation Dataset and Benchmark (BSDB)}~\cite{MartinFTM01}.
The later allow the comparison of the obtained results with other object segmentation approaches.
Each of the images has an associated ground truth in the form of a binary mask.
In the case of the two BSDS images, the ground truth masks were obtained from a previous work where 100 binary masks from objects where generated from a subset of 96 images~\cite{McGuinness:2010:CEI:1621143.1621303}.

\subsection{Windows presentation}
\label{LRSVP}

\begin{figure}
\centering
  \includegraphics[width=1.0\columnwidth ]{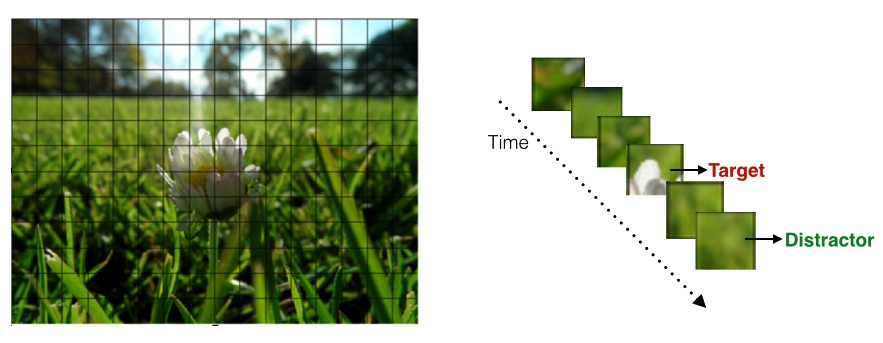}
\caption{Illustration of RSVP to randomly display different regions of an image.}
\label{fig:localRSVP}
\end{figure}

The goal of this stage is the generation of the visual stimulus in such a way that they generate a different and measurable cognitive reaction depending on whether they are associated to object or background pixels.
The approach adopted is based on the Rapid Serial Visual Presentation (RSVP)~\cite{journals/ivs/Spence02} of the different windows that compose an image containing an object of interest. 
The approach follows the same idea described in the papers for image retrieval by using BCI~\cite{Healy:2011,Wang:2009,Huang:2011} but applied at local scale. 
This involves partitioning an image into 192 windows and displaying each of them in a fast and random succession (Figure~\ref{fig:localRSVP}).
Given the homogeneous scale of the objects in the dataset and the amount of windows, these windows will usually only contain part of the object. 
In particular, the adopted ratio generated an average of 15\% of windows containing parts of the object.

A non-invasive 31 channel BCI with a sample rate of 1kHz was used to capture the brain reaction of the users during the image presentation. The electrodes were located according to the 10-20 system distribution and the experiment was run in a Faraday Cage. This room isolates the participant and equipment to minimize the interference from any other unrelated acoustic or visual events.

Image presentation in the experiments was carried out as follows. First, the entire image was displayed to the participant for five seconds. This allows the user to memorise the visual features of both object and background.
Afterwards, the 192 windows of each image were presented at a rate of 5Hz.
Each region is shown zoomed and centered on the screen. Preliminary experiments showed participants attention decreased with time. To minimise this effect, we asked participants to count the number of windows containing a part of the object.

\subsection{Preliminary experiments}
\label{mockData adquisition}

Acquiring EEG data on real users is both laborious and time consuming: in addition to the time required to actually perform the experiments (approximately one hour), it requires scheduling time with volunteers, equipment setup, and precise positioning of the various BCI sensors in a controlled environment. To ensure maximum benefit from each experiment trial, we decided to carry out a set of preliminary small-scale and simulated experiments. The objective of these experiments were: first, to establish whether classification of EEG signals with some reasonable degree of accuracy using our equipment and experiment setup is indeed feasible; second, to determine whether, given a imprecise classification of an EEG signal for a window, it is possible to use this to locate and segment the corresponding object from an image; and third, to guide us in making reasonable choices for the parameters such as the number and size of windows and their presentation rate. We include some details on these experiments here for reproducibility and to justify our design decisions. Positive results at this stage indicated that the system could indeed be effective and helped underpin the full-scale experiments.


\subsubsection{Averaging of targets and distractors}
\label{visualEEGDifference}

The first study focused on the temporal evolution of the EEG signal in those cases where this was captured at the presentation of a target or a distractor window.
Given the noisy nature of EEG signals, the observation of any difference between two individual plots from the two classes is challenging.
Nevertheless, this noise can be reduced by averaging several signals from the same class and, in this way, distinguish a clear ERP waveform.

\begin{figure}
\centering
\includegraphics[width=1.0\columnwidth ]{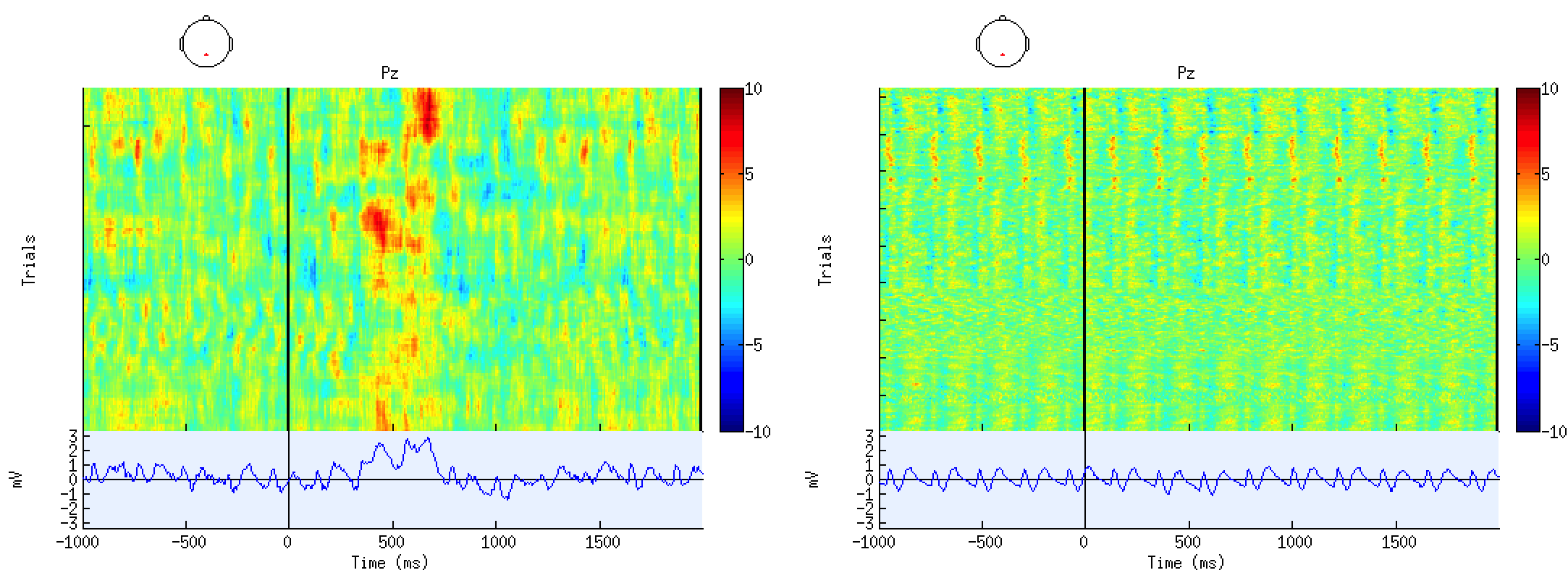}
\caption{One second before and two seconds after the visual stimulus recorded in the Pz channel for all participants (grand average). Shown are: the amplitudes of the brain waves (top), and the averaged values over all the waves for the target window (bottom-left), and distractor window epochs (bottom-right).}
  \label{fig:erptargect_distractor}
\end{figure}

Figure \ref{fig:erptargect_distractor} compares the same number of target (left) and distractor (right) signals captured in one electrode.
The time span goes from one second before the visual stimulus and two seconds after it.
The behaviour on the target reactions is different to the distractors, evidencing a peak around 500 ms after the stimulus visualization, which is clearly noticed in the averaged waveform across all the $erptargect_distractortrials$. 

This first result provided the evidence that the adopted RSVP strategy was capable of generating different and measurable brain responses for the two classes of windows.
It must be made clear that the future sections in the remainder of this paper do not apply any averaging strategy on the EEG signals associated to an image window.
All future results  presented in late sections (Section~\ref{sec:results}) are based on the classification the EEG signal obtained with a single trial.

\subsubsection{Binary classification and simulated EEG maps}
\label{firstSVMandSimulatedEEGmaps}

A second test was performed to establish the feasibility of distinguishing between target and distractor windows using EEG signals. We posed this as a binary classification problem and trained a binary SVM with RBF kernel classifier with target and distractor EEG signals. 459 EEG signals were used to train the classifier (229 targets and 230 distractors), and 153 EEG signals for testing (76 targets and 77 distractors). The zero-one accuracy obtained was 0.68, which shows sufficient signal is present to achieve better than random classification.

\begin{figure}
\centering
\includegraphics[width=1.0\columnwidth ]{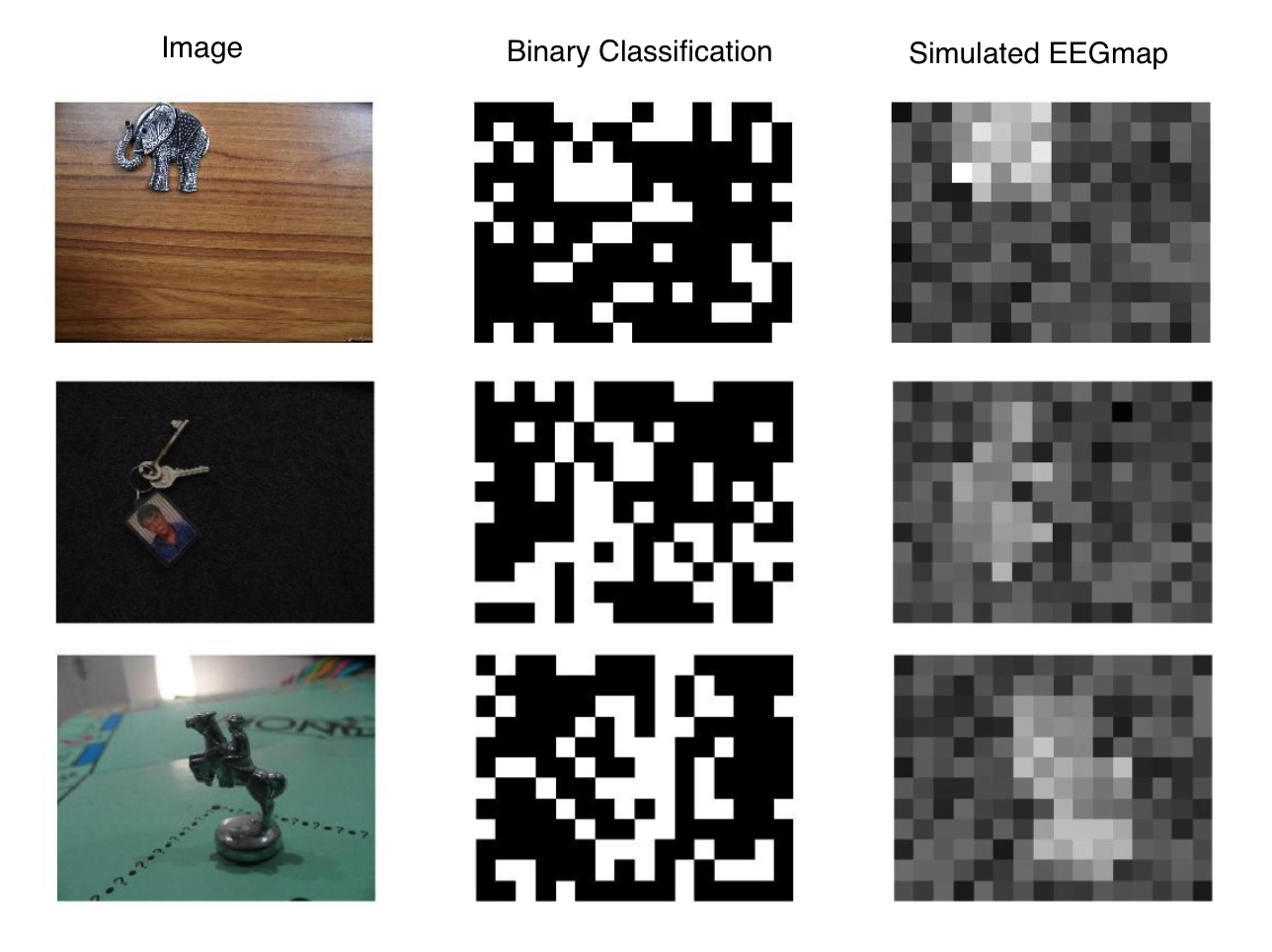}
\caption{Left: original image; centre: simulation of the window labels (white: windows with object, black: window with background); right: probability map from simulating the SVM scores.}
  \label{fig:simulatedEEG map}
\end{figure}

This final preliminary experiment was intended to determine if, given a noisy classification signal from an SVM trained on EEG signals, this could be used to seed a segmentation. We simulated the output of a binary classifier on ground truth images using draws from a Bernoulli distribution with $P(X=1) = 0.68$ for windows containing a target. Figure~\ref{fig:simulatedEEG map} (center) illustrates the resulting binary classification maps. The results are, clearly, quite noisy; significant information is lost when the SVM scores are binarized. We therefore chose to use the normalized SVM scores, rather than thresholded decisions, to estimate a soft probability. To simulate SVM scores, we model the distribution of scores given the classification decision as Gaussian, fitting parameters from the data used in the first preliminary experiment, and draw from these Gaussians conditioned on the binary classification decision.  Figure~\ref{fig:simulatedEEG map} (right) shows the resulting generated probability maps, which clearly highlight the object of interest. We chose conditioned Gaussians based on histogram observations; note, however, that this assumption has no bearing on the remainder of the experiments. The simulation results indicate that SVM scores are a useful estimator of the probability that a particular region contains a target.


\section{EEG processing}
\label{sec:eegprocessing}

In this section we describe the actual procedure (i.e. based on what was learnt from the preliminary experiments reported in the previous section) followed to clean and classify the brain signals related to the windows presented to the users. 
The output generated in this stage are the EEG maps that will be used to produce the final object segmentation for the images.

\subsection{Data cleaning and feature vectors}
\label{subsec:datacleaning}

The data was referenced to the Tp9 channel and subsampled from the original 1000Hz rate to 250Hz. For 3 of the users the Tp10 channel was used instead due it was cleaner and, therefore, introduced less noise to the raw signals of the rest of the channels. Then, a band-pass filter from 0.1Hz to 70Hz was applied. By visual inspection, we rejected manually the noisy segments. With the data filtered, we extracted the brain reaction related to the stimulus by selecting one and two seconds pre and post-window presentation (epochs).

For the feature selection, we selected the time region within the epoch that best characterizedd the difference between targets and distractors. As shown in Figure~\ref{fig:erptargect_distractor}, this region is contained between 200ms and 900ms after the visual presentation. The feature vectors are built by concatenating the 31 channels for this time region. The final feature vector is obtained by applying a second subsample to the vectors to reduce the sample rate to 20Hz.

\subsection{Binary classification of windows}
\label{subsec:classification}

We worked with the scikit-learn Python library~\cite{scikit-learn} to train the SVM with RBF kernel classifier. 
The feature vectors were normalized with zero mean and unit standard deviation across each feature component. From the total amount of 22 images, 17 were selected to train the classifier. The EEG vectors related to these images formed an imbalanced set of 435 examples of targets and 2829 examples of distractors, respectively labeled with 1 and 0. An SVM with RBF kernel was trained, and grid search with 5-fold cross validation was used for hyperparameter selection. The parameters selected were the ones that obtained the maximum averaged Area Under the Curve value (AUC) across all the folds. 

The final model was tested on 5 images, which contained a set of 130 targets and 830 distractors. Table~\ref{table:accuracy_It1} gives the measured performance.

\begin{table}[t] 
\caption{Area Under the Curve (AUC) and Averaged Precision (AP) obtained per user\vspace{0.5em}} 
\centering 
\setlength{\tabcolsep}{8.5pt}

\begin{tabular*}{\columnwidth}{l r r r r r r r}  
  \toprule
User &     1 &     2 &     3 &     4 &     5 &    avg &   std \\ \midrule
AUC  &   .63 & .75 & .73 & .78 & .65 & .71  & .06 \\
AP  &    .22 & .33 & .30 & .45 & .23 & .31   & .08 \\

  \bottomrule
\end{tabular*} 
\label{table:accuracy_It1}
\end{table}

\begin{figure}
\centering
\includegraphics[width=1.0\columnwidth ]{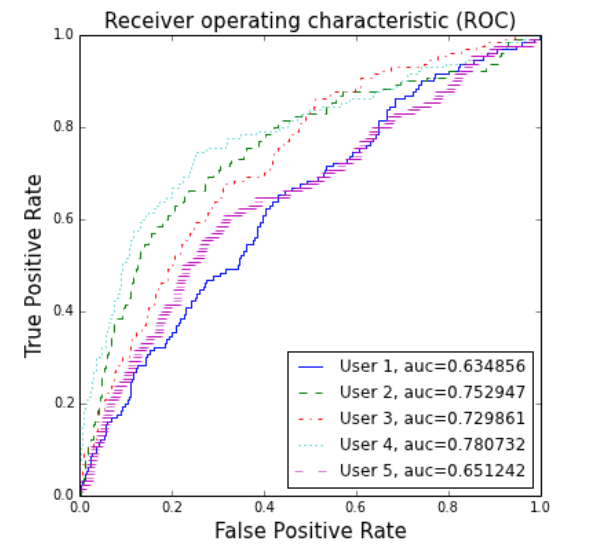}
\caption{Classification performance per each user in the Reciever Operating Characteristic (ROC) space.}
  \label{fig:ROC}
\end{figure}

\subsection{EEG maps}
\label{subsec:EEG maps}
The confidence scores provided by the classifier can be graphically represented as an image in the form of \emph{EEG maps}.
This score represents the distance that separates the classified sample from the hyperplane~\cite{scikit-learn}. 
Depending on the sign of this distance, the binary classifier assigns a target or distractor label.
The maps are built by normalizing the values assigned to each window between 0 and 1 according to: 
\begin{equation}
X'=\frac{X-min(X)}{max( X)-min(X) },
\label{eq:normalization}
\end{equation}
where $X'$ represents the EEG map normalized and $X$ the original EEG map. 


\begin{figure}
\centering
\includegraphics[width=1.0\columnwidth ]{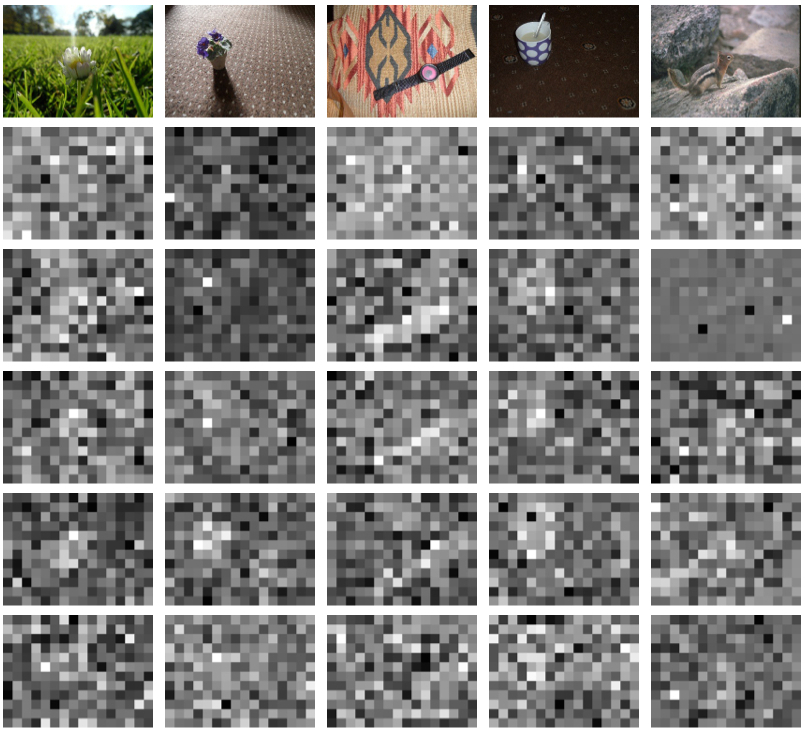}
\caption{EEG maps for a test set of images. The top row is the original images and the remaining rows are the generated EEG maps for five different participants. Brighter pixels represent higher probabilities.}
  \label{fig:EEG mapS}
\end{figure}

\section{Object segmentation}
\label{sec:segmentation}

The EEG maps constructed in the previous section provide local information about how likely is to find an object part in each window. 
The final segmentation requires a post-processing of the EEG maps to obtain a pixel-wise binary mask of the object location.
Three configurations have been assessed for this task, and for each it was required learn a different set of parameters:
\begin{enumerate}
\item Binarization of the EEG maps
\item Filtering and binarization of the EEG maps
\item Filtering and binarization of the EEG maps to seed a segmentation algorithm.
\end{enumerate}

EEG maps are generated after training the SVM model on 17 images. 
The different values for the segmentation parameters were learned on these training images based on the average performance of the 17 processed EEG maps. 

The quality of the segmentation was evaluated with the Jaccard Similarity Index, a popular metric for object segmentation used, for example, in Pascal Visual Object Classes (VOC) Challenge~\cite{Everingham10}.
This measure is made to evaluate the similarity between the final segmentation and ground truth masks. The Jaccard Index has values between 0 an 1, with 1 the maximum similarity between the masks. The measure is defined as the intersection of the two final binary masks divided by the union of both masks:
\begin{equation}
J(A,B) =\frac{A \cap B}{A \cup B},
\end{equation}
\label{eq:jaccard}
where $A$ is the segmentation mask and $B$ is the ground truth mask.

\subsection{Binarizing the EEG maps}
\label{subsec:binarizingEEG maps}

The simplest strategy to quantitatively assess the EEG maps in terms of object localization is to directly convert them into a binary mask.
Such binarization is achieved by setting a threshold $\alpha$, which will consider as targets all those pixels in the EEG map which are higher than $\alpha$, and label as distractors all the rest.
An optimal binarization threshold $\alpha_i$ was estimated for each individual user $i$ by averaging the $\alpha_{i,j}$ values that provided the highest Jaccard index for each training image $I_j$.

\begin{equation}
\alpha_{i,j} = \underset{\alpha}{\operatorname{arg min}} \: J(M_{i,j}(\alpha), GT_j)
\label{eq:bestJaccard}
\end{equation}
where $M_{i,j}$ is the EEG map thresholded by $\alpha$ for user $i$ and image $I_j$, and $GT_j$ is the ground truth mask for image $I_j$. 

\begin{figure}[t]
  \centering
  \includegraphics[width=1.0\columnwidth ]{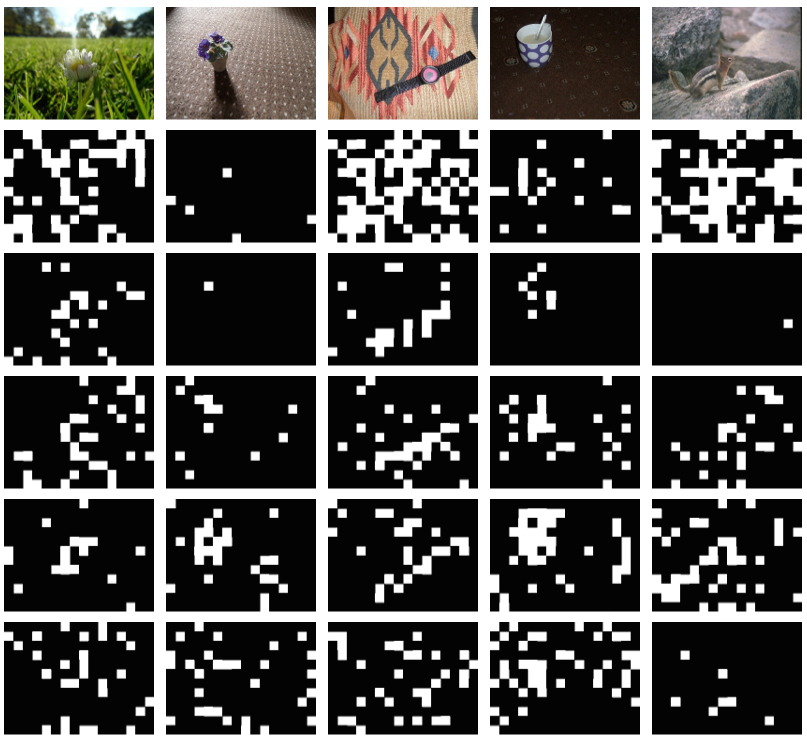}
  \caption{Binary masks after thresholding the EEG maps. The top row is the original images and the remaining rows are the binary masks for five different participants.
}
  \label{fig:ConfA_BinaryUsers}
\end{figure}

Quantitative results for this approach are presented in Figure~\ref{fig:ConfA_BinaryUsers} have a high density set of windows labelled as target around the object location, especially for user 4. 

Table~\ref{table:confA} contains the thresholds learned for each of the six users in the test by using 17 images for training.
The table also includes the Jaccard index for each user when these thresholds are applying on the 5 test images.
The averaged Jaccard index through all the users corresponds to a low $0.14$, which points at the poor performance of a direct binarization on the EEG map.


\begin{table}[ht] 
\caption{Final threshold per user obtained from the EEG maps for training and the final average value obtained applying the threshold on the test set.\vspace{0.5em}}
\centering 
\setlength{\tabcolsep}{8.5pt}
\begin{tabular*}{\linewidth}{lrrrrrrr} 
\toprule
User      &   1 &   2 &   3 &   4 &   5 &   avg & std \\ \midrule
$\alpha$  & .59 & .74 & .65 & .61 & .67 & .65 & .06 \\
$J$    & .10 & .15 & .18 & .21 & .11 & .23 &  .17 \\
\bottomrule
\end{tabular*} 
\label{table:confA} 
\end{table} 

\subsection{Filtering and binarization of EEG maps}
\label{subsec:filteringSigmaEEG}

The binarization approach presented in the previous section presents a first limitation because of the block artefacts introduced by the window boundaries.
The window contours do not need to match with the object ones, so in general this lack of resolution is partially responsible of the bad performance of the solution.
In addition, the spatial relationship between the windows is completely ignored, without any contextual analysis that may provide coherence to the overall composition.


In this section, a low-pass filter is added before thresholding the maps to reduce block artefacts. 
With this filter, the isolated false positive windows of the background can be reduced and the high compact windows around the object will mutually reinforce.
Equation~\eqref{eq:exprGauss} describes the filter mask (kernel) that is convoluted with the image. 
The $(x,y)$ values are the horizontal and vertical distances from the origin to a certain point of the kernel. 
The kernel takes standard deviation $\sigma$ as a parameter defining the spatial extension of the filter:
\begin{equation}
G(x,y) = \frac{1}{\sqrt[]{2\pi\sigma^2}}\exp{-\frac{x^2+y^2}{2\sigma^2}}
\label{eq:exprGauss}
\end{equation}

The Gaussian filtering and posterior binarization of the resulting EEG map requires defining the two parameters $\alpha$ and $\sigma$.
As in the previous section, these were selected via minimizing error the training dataset.
In this case, though, the Gaussian filtering changes the dynamic range of the EEG maps the threshold, which is no longer between 0 and 1.
For this reason, the binarization threshold is not learnt as an absolute value but as a normalised coefficient $p\in [0,1]$ referred to the dynamic range of the EEG map:
%
%
\begin{equation}
\alpha_{i,j}(p) = min(F_{i,j}) + p\cdot(max(F_{i,j})-min(F_{i,j})) ,
\label{eq:percentage}
\end{equation}
where $F_{i,j}$ the filtered EEG map of user $i$ for image $I_j$.


The procedure used for optimisation was to select the parameters ($\sigma, \alpha$) that generated the maximum averaged Jaccard Index over all the images of the train set. 70 values for sigma ($\sigma\in [0, 70]$) were tested to filter the EEG map. For each filtered map, 100 different values were tried by varying $p$ from 0 to 1, and the binarization threshold $\alpha_{i,j}$ that maximized the Jaccard was selected, as previously presented in Equation~\eqref{eq:bestJaccard}. Then, for each image a optimal combination ($\sigma, \alpha$) that maximized the Jaccard was obtained.
Finally, the parameters used in the test set were set by averaging the 17 pairs of optimal parameters computed for the training set.



\begin{figure}[t]
  \centering
  \includegraphics[width=1.0\columnwidth ]{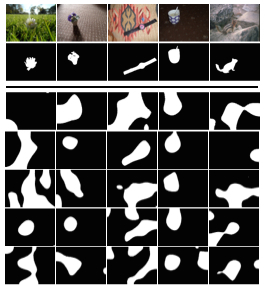}
  \caption{Binary mask after filtering and thresholding the EEG maps. Each row represent the final masks per user.
}
  \label{fig:ConfB_BinaryUsers}
\end{figure}

The new binary masks shown in Figure~\ref{fig:ConfB_BinaryUsers} present in many cases a single patch located near the actual position of the object, with a shape which is much more natural than the sparse blocks generated in Figure~\ref{fig:ConfA_BinaryUsers}.
A quantitative analysis of the results is presented in Table~\ref{table:confB} and results in an important gain of 79\% when comparing the averaged Jaccard indices of thresholding with our without the Gaussian filtering. 
The results combining a low-pass filter with thresholding the EEG maps produce a cleaner binary masks which contain a better estimation for the object location (Figure~\ref{fig:ConfB_BinaryUsers}). 
However, these values are still too poor to consider these results an accurate segmentation of the object.

\begin{table}[t] 
\caption{Averaged percentage (normalized to one) and $\sigma$ per user obtained from the train set and final Jaccard index obtained on the test set by using these parameters.\vspace{0.5em}}
\centering

\setlength{\tabcolsep}{4.7pt}
\begin{tabular*}{\linewidth}{lrrrrrrr} 
\toprule
User      &     1 &     2 &     3 &     4 &     5 &      avg &  std \\
\midrule
$p$       &   .61 &   .70 &   .63 &   .65 &   .69 &    .66 &  .04 \\
$\sigma$  & 39.24 & 33.29 & 26.35 & 29.18 & 33.53 & 32.32 & 4.89 \\
J      &   .16 &   .30 &   .27 &   .41 &   .24 &   .27 &  .09 \\
\bottomrule
\end{tabular*} 
\label{table:confB}
\end{table} 

\subsection{Seeding the segmentation algorithm}
\label{subsec:filtersigmagrabcut}

The results obtained in the previous section, based only on EEG data, already provide in many cases a rough estimation of the object location.
The configuration explored in this section explores the synergy between BCI data and computer vision algorithms.
The EEG maps filtered with a Gaussian kernel are used to seed an object segmentation algorithm that can exploit the spatial dependencies between neighbouring pixels.
This way the computer vision algorithm is guided by the user in a noisy and approximate fashion.

The segmentation algorithm used is GrabCut~\cite{Rother:2004:GIF:1015706.1015720}. 
This technique performs a segmentation of an image based on a rough initial segmentation defined by the user, typically by drawing a box around the target object. 
The pixels outside the box are initially considered as background and the pixels inside as unknown. 
The technique models separately the pixels labeled as background and the ones labeled as unknown by using a Gaussian Mixture Model (GMM). 
The unknown pixels are considered foreground pixels in the first iteration. Then, the two GMMs obtained are used to solve a minimization problem via min-cut and produce a first segmentation of the object. 
After the initial iteration, with the new labels for background and foreground, GMM are updated and the process is repeated until converge on the final segmentation.
Our proposal here is to replace the drawn rectangle by using the EEG maps. 

The popular OpenCV~\cite{opencv_library} implementation of GrabCut was selected for this purpose. 
The algorithm requires an input a map with pixels marked as:
\begin{inparaenum}[\itshape a\upshape)]
\item definitely background;
\item possible background;
\item possible foreground; and
\item definitely foreground.
\end{inparaenum}
The algorithm requires labels \emph{a} and \emph{b} or \emph{c}. Label \emph{d} is optional, and we have not considered to assign this value due to the noisy nature of the signals.


The initialization of GrabCut with EEG maps requires thresholding the Gaussian filtered map by applying two thresholds: $\alpha_{1}$ to separate pixels (a) from the rest, and $\alpha_{2}$ to separate the pixels labeled as (b) from pixels of (c). 
Both thresholds are defined as relative percentages ($p_{1}, p_{2}$), as in the Section~\ref{subsec:filteringSigmaEEG} eq.~\eqref{eq:percentage} for the same reason: after applying the low pass filter, the EEG map is unnormalized.

The optimization is realized by randomized search~\cite{Bergstra:2012:RSH:2188385.2188395}, trying combinations of ($p_{1}, p_{2}, \sigma$) and computing the final Jaccard Index. We used the hyperopt python package~\cite{BergstraYC13} for the optimization problem.  

The function to optimize (eq.~\eqref{eq:optFunct}) is the one that computes the average of all the Jaccard Index for the training set given a parameters combination. The optimization is to find the parameters that minimize the error on the averaged Jaccard index:
\begin{equation}
E( p_{1}, p_{2}, \sigma ) = 1 - \frac{\sum_{j=1}^{N} J(M_{i,j}(p_{1}, p_{2}, \sigma), GT_{j})}{N},
\label{eq:optFunct}
\end{equation}
where $N=17$ is the number of training images. First, the EEG maps are filtered by applying a Gaussian filter with the sigma parameter $\sigma \in [0,70]$, then the filtered map is thresholded at two levels by applying two thresholds: $p_{1} \in [0, 0.5) $ and $ p_{2}\in [0.5, 1.0]$. We randomly pick 1,000 combinations.

\begin{table}[h] 
\caption{Optimal parameters ($p_{1}$, $p_{2}$, $\sigma$) per user. $p_{1}, p_{2}$ are the percentages normalized to 1. Accuracy is the final Jaccard index on the test set.\vspace{0.5em}} 
\centering
\setlength{\tabcolsep}{4.7pt}
\begin{tabular*}{\linewidth}{lrrrrrrr} 
\toprule
User     &     1 &     2 &     3 &     4 &     5 &      avg &   std \\
\midrule
$t_1$    &   .04 &   .15 &   0.02 &   .27 &   .17 &   .17 &   .18 \\
$t_2$    &   .77 &   .76 &   .64 &   .81 &   .69 &   .73 &   .07  \\
$\sigma$ & 35.65 & 29.56 & 37.53 &  3.18 & 19.10 & 25.00 & 14.16 \\
Jacc.    &   .28 &   .62 &   .31 &   .36 &   .69 &   .45 &   .19  \\
\bottomrule
\end{tabular*} 
\label{table:confp}
\end{table} 

\begin{figure}[t]
  \centering
  \includegraphics[width=1.0\columnwidth ]{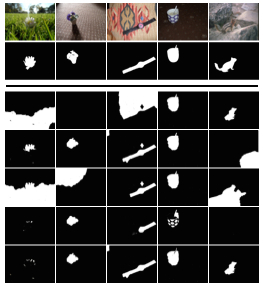}
  \caption{Final binary segmentations. The images and the ground truth masks are displayed (first an second row). Below, each row represents a different user.}
  \label{fig:ConfC_binary}
\end{figure}

\section{Results}
\label{sec:results}

\begin{figure}[t]
  \centering
  \includegraphics[width=1.0\columnwidth ]{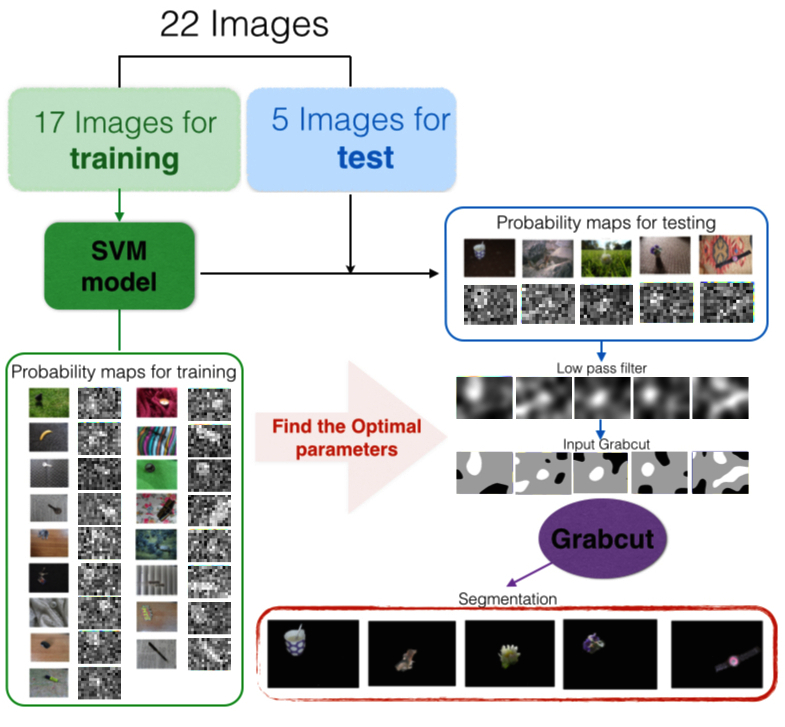}
  \caption{Diagram of the segmentation procedure. The training set is used to generate the SVM model to produce the EEG maps. The same SVM is used on the training images to produce a EEG maps for training, from where the system parameters are estimated. The separated test images ares classified by the SVM and the final segmentation is performed by using the learned parameters.
}
  \label{fig:crossValDiagram}
\end{figure}

As the number of images for testing the system is limited, a cross-validation is performed by switching the images on the test and training set 5 times and producing that way the segmentation of all the dataset. That means that 5 different systems are generated following the pipeline described (Figure \ref{fig:crossValDiagram}), where the 5 testing images are always independent from the training.

\begin{figure}[t]
  \centering
  \includegraphics[width=1.0\columnwidth ]{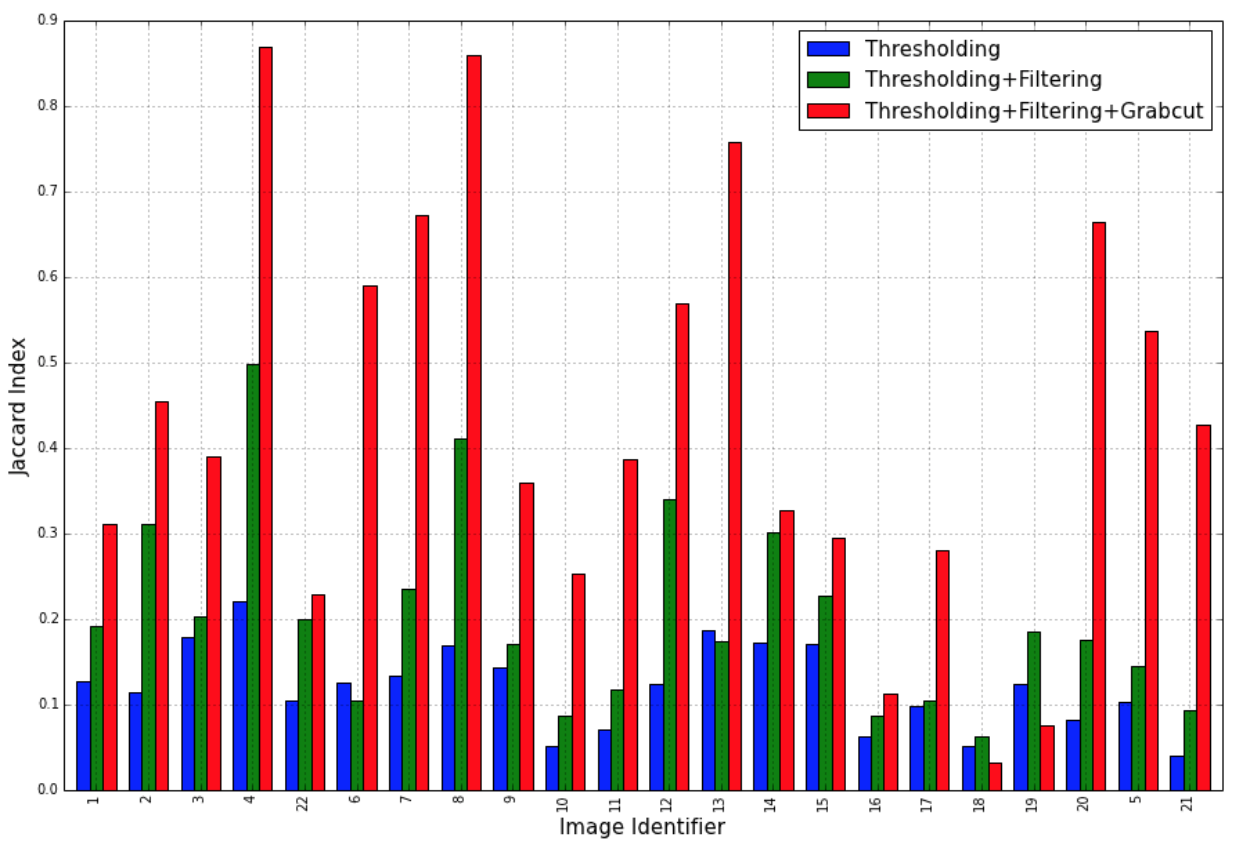}
  \caption{Average Jaccard index across users per image.}
  \label{fig:JaccardImage}
\end{figure}

The results obtained are plotted in Figure~\ref{fig:JaccardImage} for the three strategies to produce the final binary mask introduced in Section~\ref{sec:segmentation}.
Averaged Jaccard accuracies indicate that the configuration of using GrabCut with the filtered and thresholded EEG maps performs better than the other configurations, producing a good binary mask in many of the images. However, in other images, the Jaccard Index is not high enough and the segmentation is noisy.

\begin{figure}[t]
  \centering
  \includegraphics[width=1.0\columnwidth ]{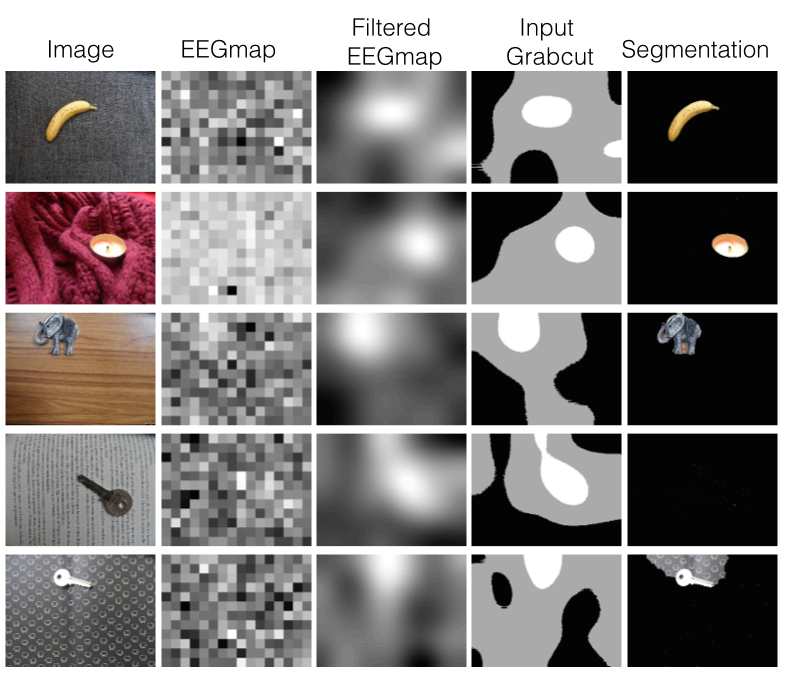}
  \caption{All steps to produce the segmentation combining the EEG maps with GrabCut. Results are from the user 2; segmented images are the set used on the second iteration of the cross-validation. 
}
  \label{fig:exampleSegmentation}
\end{figure}

Figure~\ref{fig:exampleSegmentation} presents the visual segmentation for five examples, as well as the intermediate stages.
The first three results offer a good qualitative segmentation, while the two last do not succeed in the task.
These two failing examples share the characteristic of a very similar distribution between the object and the background.
While the EEG map offers a reasonable quality, the GrabCut algorithm fails in the segmentation.
This effect is possibly due to the color-driven approach adopted by GrabCut, which basically models foreground and background with color GMM.

It is possible to see that the filtered EEG maps produce a good estimation of the object location that in three of the five examples produce a good segmentation. The segmentation is less accurate in two images, although the location in the processed EEG map is reasonable.


These results show that it is possible to successfully classify the brain reaction produced to detect different parts of a target object, and to produce useful information based on the EEG waves to locate the target object on the images.

\begin{table}[t] 
\caption{Final Jaccard for each of the five iterations of the cross-validation of the system\vspace{0.5em}}
\centering 
\setlength{\tabcolsep}{10pt}
\begin{tabular*}{\linewidth}{lrrr}
\toprule
       & Jaccard A & Jaccard B & Jaccard C   \\
\midrule
avg & .13 & .21 & .47 \\
std  & .03 & .04 & .12 \\

\bottomrule
\end{tabular*} 
\label{table:crossValidation} 
\end{table}  

\subsection{Combining EEG maps of different users}
\label{subsec:EEGmapsComb}

To reduce the noise of the EEG maps, we compute an unique map per image by averaging the EEG maps of the different users. The final segmentation is performed following the approach described in the section \ref{subsec:filtersigmagrabcut}. The parameters are picked by averaging across iterations and users ($p_{1}=0.23$, $p_{2}=0.74$, $\sigma=24.10$). 

\begin{figure}[h!]
\centering
\includegraphics[width=1.0\columnwidth ]{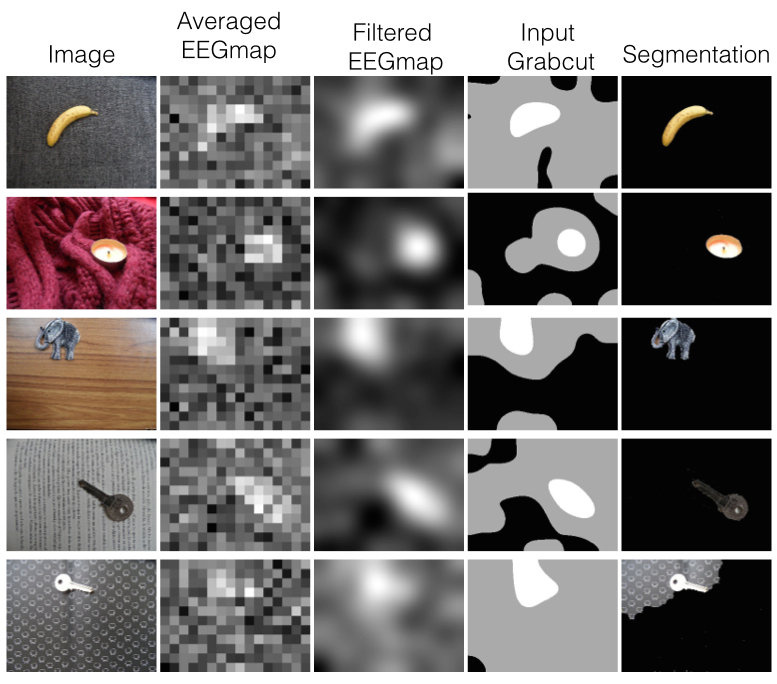}
\caption{All steps to produce the segmentation combining the EEG maps with GrabCut. The EEG maps are the average of the all the user's maps generated. 
}
\label{fig:averagedEEGmaps_}
\end{figure}

Qualitative results of the averaged EEG maps provide evidence that combining the individual maps of different users it is possible to generate cleaner EEG maps (Figure \ref{fig:averagedEEGmaps_}). The final Jaccard combining the EEG maps of the users outperform in 18 of the 22 images, getting an averaged Jaccard of 0.72, 1.6 times superior to the global result obtained before (Table \ref{fig:averagedEEGmaps_}).

\begin{figure}[h!]
  \centering
  \includegraphics[width=1.0\columnwidth ]{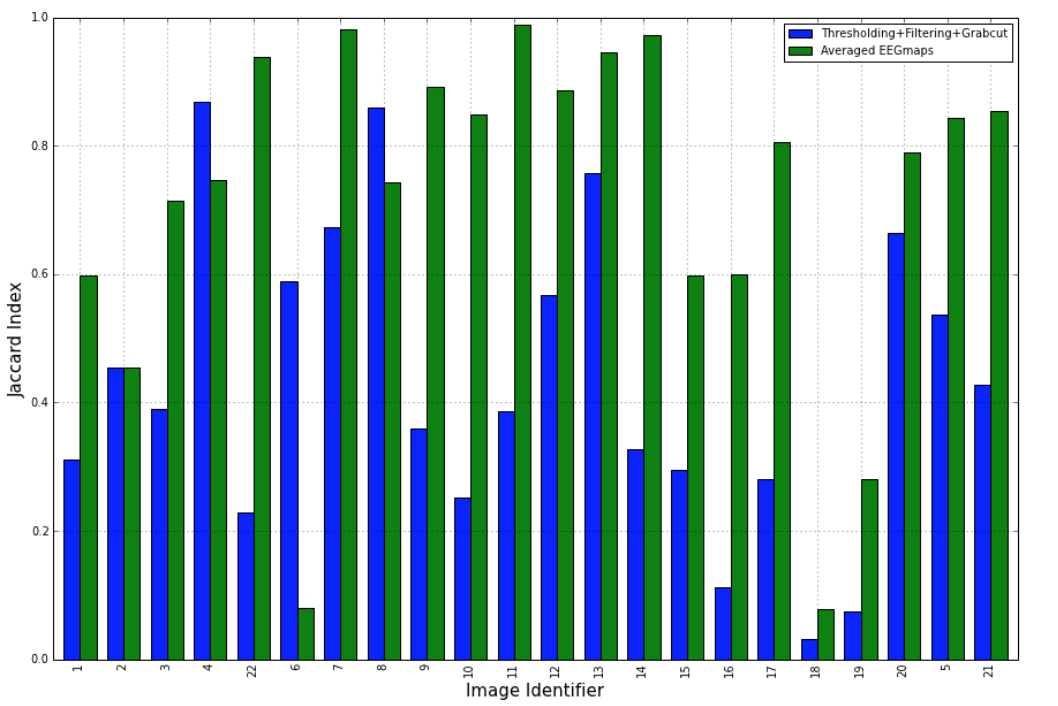}
  \caption{Final Jaccard per image. In blue, EEG maps combined with GrabCut. In green, averaged EEG maps across users combined with GrabCut. 
}
  \label{fig:averagedEEGmaps}
\end{figure}

\subsection{EEG vs. mouse-based interfaces}
\label{berkeley}

The analysis of the proposed BCI-based solution is compared with a state-of-the-art solution using a mouse instead of the human-computer interface.
The study is based on the two images from the Berkeley Segmentation DataSet (BSDS)~\cite{MartinFTM01} we used, which were also considered in a previous study on mouse-based segmentation tools~\cite{McGuinness:2010:CEI:1621143.1621303}. 
In that work, four different segmentation techniques were compared in an interactive set up where users draw scribbles to seed the algorithm.
That experimentation measured the evolution of Jaccard index with respect to the amount of time that a user was engaged in the operation of the tool.

For the sake of a fair comparison, only the image processing algorithm referred as Interactive Graph Cuts (IGC) has been considered because it is the most similar to the GrabCut solution adopted in our paper.
In terms of time evolution, we selected the Jaccard index obtained 45 seconds after the start of the interactive segmentation because this is the closest time stamp to the total display time of an image in our EEG-based system: 43.4 seconds distributed in an initial display of the full image during 5 seconds and 38.4 seconds for the RSVP of 192 windows at a rate of 5 Hz.



The average Jaccard indexes for the EEG- and mouse-based segmentations are presented in the Table \ref{table:berkeley}.
The obtained figures clearly show that mouse-based interaction outperforms the proposed EEG-based method which, in addition, requires the costly task of installing the BCI on the user. Note that the high variability on the standard deviation associated to the averaged Jaccard evidence a high variability in the users performance for the EEG results, different to the mouse-based interface, where all the users perform similar.
%
\begin{table}[ht] 
\caption{Jaccard Index for EEG and mouse-based interaction methods.\vspace{0.5em}} 
\centering 
\setlength{\tabcolsep}{20pt}

\begin{tabular*}{\columnwidth}{l c c c }  
  \toprule
            &     BSDS 38082 &   BSDS 123074 \\ \midrule
   EEG      &    $.36 \pm .33$ &   $.23 \pm .23$ \\
     Mouse  &    $.74 \pm .06$ &   $.89 \pm .02$ \\
  \bottomrule
\end{tabular*} 
\label{table:berkeley}
\end{table}


%

\section{Conclusions and future work}
\label{sec:conclusionsandfuturework}

We proposed a system for object segmentation using brain signals. The system is posed as a proof of concept, with the objective being to determine if such a system is feasible. 

We designed a specific method of presenting images to associate each image region with its visual brain reaction. Our use of non-overlapping blocks limits the resolution of the generated EEG maps; future work will consider overlapping windows increase the spatial resolution.

The EEG processing performed in the paper (Section~\ref{sec:eegprocessing}) is based on low-pass filtering the EEGdata, epoching the data to identify each window with its brain reaction, training an SVM with the the down-sampled signal, and concatenating EEG channels to form a feature vector. This simple processing gives an AUC of .71; more sophisticated analysis, e.g. Independent Component Analysis (ICA) for channel selection and artifact removal, may improve classifier performance. 
Better EEG features, such as wavelet features, may also improve classification and, consequently, the quality of the EEG maps.

In Section~\ref{sec:segmentation} we investigated three different configurations to produce binary masks. The EEG maps obtained are noisy and require post-processing; a Gaussian low-pass filter was effective in reducing the effect of noise and improves Jaccard accuracy (\ref{subsec:filteringSigmaEEG}). Using other filters or morphological operators on the binarized EEG map may improve results. Subsection~\ref{subsec:filtersigmagrabcut} discusses our preprocessing of the EEG maps to set the initial inputs to GrabCut. 
Future work will consider using the values of the EEG maps directly to set initial terminal capacities of the min-cut graph.

We have shown that the fusion of different user's EEG maps helps to reduce the noise on the probability masks and perform a segmentation that, once combined with GrabCut, notably outperforms the segmentation performance acquired.

Our system shows that it is possible to roughly locate and delineate an object in an image using EEG data, but is far from get the quality on the segmentation of other state-of the art interactive segmentation tools.
Nonetheless, this proof of concept opens the door to new interaction modes which may become specially valuable for those people affected by Locked in Syndrome.
For them, this work may represent a promising direction in improving their communication for applications such as object selection.
If the accuracy of BCI keeps increasing, and their cost decreasing, it is expected that new applications will appear for this novel human computer interface that has raised the interest in different fields of the multimedia community.

\section{Acknowledgments}
This publication has emanated from research conducted with the financial support of Science Foundation Ireland (SFI) under grant number SFI/12/RC/2289 and partially funded by the Project TEC2013-43935-R BigGraph of the Spanish Government.

%
\bibliographystyle{abbrv}
\bibliography{acmmm-hrhr-2014_arxiv}  
%
%

\end{document}